\documentclass[11pt]{article}
\usepackage[margin=1in]{geometry}
\usepackage{graphicx}
\usepackage{booktabs}
\usepackage{amsmath}
\usepackage{hyperref}
\usepackage{natbib}
\usepackage{xcolor}
\usepackage{multirow}
\usepackage{caption}
\usepackage{subcaption}
\usepackage{enumitem}
\usepackage{colortbl}
\usepackage{float}
\graphicspath{{figures/}}

\title{Stance Detection in Prediction Markets:\\
Addressing Imbalanced Trader Commentary via\\
Counterfactual Augmentation and Market Context}
\author{
  Thomas Mbrice\\
  Department of Computer Science\\
  Stony Brook University\\
  \texttt{thomas.mbrice@stonybrook.edu}
  \thanks{Code, data, and trained model checkpoints are available at \url{https://drive.google.com/drive/folders/1K8LU8Q0UXjljJhtjShbOIwkH\_l74sD1m?usp=sharing}.}
}
\date{May 2026}

\begin{document}
\maketitle

\begin{abstract}
Prediction markets such as Polymarket aggregate crowd beliefs into real-time probability estimates, and the comments traders post beneath each market contain rich directional stance signals that prices alone cannot capture. This work introduces the first stance detection study applied to prediction market commentary, a domain characterized by extreme brevity, trader-specific vernacular, and severe class imbalance (only 8.7\% of comments oppose the market outcome). RoBERTa-base is fine-tuned across a $4 \times 3$ ablation: four input configurations (\{2-class, 3-class\} $\times$ \{with/without market context\}) and three augmentation conditions (baseline, 50\% synthetic, 100\% synthetic). Synthetic minority-class samples are generated via LLM-driven Pro$\to$Anti counterfactual flips using the Anthropic API. Results show that (1) market context is the single most impactful factor, raising 3-class \textsc{Anti} recall from 0.10 to 0.45; (2) counterfactual augmentation is conditionally effective, improving \textsc{Anti} F1 in weak configurations (0.10 $\to$ 0.24) while degrading strong ones (2-class-ctx macro F1: 0.68 $\to$ 0.50 at full dose); and (3) 50\% augmentation is the optimal dose, with 100\% consistently hurting performance. Attention-based interpretability analysis provides mechanistic support for all three findings.
\end{abstract}

\section{Introduction}

Prediction markets aggregate crowd and popular beliefs into real-time probability estimates for future events. Platforms like Polymarket have attracted significant attention from both retail and institutional participants, positioning market odds as sentiment indicators alongside traditional financial data feeds, with characteristics unique to social media platforms. While market prices reflect aggregate trader beliefs, the \textit{comments} posted beneath each market contain rich directional stance signals that prices alone cannot capture.

Extracting these signals is challenging for three compounding reasons. First, prediction market text is extremely short, often a single phrase or sentence. Second, comments are primarily written in trader-specific slang (e.g., ``this is cooked'', ``omg'', ``free money'') that diverges sharply from the political and journalistic text on which most stance models are trained. Third, the data exhibits severe class imbalance: only 8.7\% of comments express opposition to the market outcome (\textsc{Anti\_Position}), making standard accuracy-based evaluation misleading and requiring targeted minority-class strategies.

This paper addresses these gaps through three contributions. First, a 2,229-comment annotated dataset is constructed and released spanning 12 Polymarket markets across politics, sports, and finance, the first such resource for prediction market stance detection. Second, a market context input transformation is proposed and ablated, prepending the market question to each comment and providing zero-cost grounding at inference time. Third, an LLM-driven counterfactual augmentation pipeline is designed that flips \textsc{Pro\_Position} comments to \textsc{Anti\_Position} while preserving trader tone, and the dose-response relationship of this augmentation is systematically evaluated.

The experimental design is a full $4 \times 3$ ablation: 4 model configurations $\times$ 3 augmentation conditions, yielding 12 experimental runs evaluated on a fixed stratified test set using macro F1. Additional analyses include attention map visualization, confusion matrix comparison, dose-response plotting, and per-market performance breakdown.

The remainder of this paper is organized as follows. Section~\ref{sec:related} reviews related work on stance detection and counterfactual augmentation. Section~\ref{sec:method} describes the three methodological contributions. Section~\ref{sec:setup} details the experimental setup. Sections~\ref{sec:results} and \ref{sec:analysis} present results and analysis. Section~\ref{sec:conclusion} concludes.

\section{Related Work}
\label{sec:related}

\paragraph{Stance Detection.}
Stance detection has been studied extensively on political tweets \citep{mohammad2016semeval, li2021pstance} and social media debates \citep{allaway2020vast}, typically employing transformer-based classifiers such as BERT \citep{devlin2019bert} and RoBERTa \citep{liu2019roberta}. More recent work has examined robustness to topic shift \citep{kim2025stancec3} and pretrained-model bias \citep{zhang2024rccl}. However, no prior work addresses stance in prediction market text, which presents distinct properties: direct financial risk attached to expressed beliefs, extreme comment brevity, and trader-specific vernacular absent from standard NLP corpora. Furthermore, no annotated datasets for this task previously existed.

\paragraph{Counterfactual Data Augmentation.}
Counterfactual data augmentation (CDA) was pioneered by \citet{kaushik2020learning}, who showed that minimal edits to input text that flip the label can reduce model reliance on spurious surface features. \citet{kaushik2021explaining} subsequently provided theoretical grounding for why such edits are effective. CDA has been applied to stance detection by \citet{li2025factual}, who used LLM-generated counterfactuals to calibrate pretrained model bias, and by \citet{zhang2024rccl}, who proposed contrastive learning over counterfactual pairs. \citet{nakada2026synthetic} provide a theoretical analysis of LLM-based synthetic oversampling more broadly.

\paragraph{Dose Sensitivity and Failure Modes.}
The augmentation literature consistently reports diminishing or negative returns beyond a domain-dependent dose threshold. \citet{karimi2023caisa} found that even 400 augmented samples introduced noise in causal claim identification; \citet{joshi2022investigation} documented settings where CDA is ineffective or harmful. The present work extends these findings to the prediction market domain, where the optimal dose depends critically on baseline model strength.

\section{Methodology}
\label{sec:method}

Three methodological contributions are proposed to address the challenges of prediction market stance detection.

\subsection{Market Context as Input Transformation}

Prediction market comments are frequently ambiguous in isolation. Phrases like ``this is moving'' or ``rip'' carry no directional stance signal without knowing what market outcome is being discussed. Prepending the market question provides grounding at zero additional annotation cost.

\paragraph{Motivation.} The model needs to know \textit{what} is being bet on to interpret short, slang-heavy comments. Without context, a comment like ``Yoo Seong moving'' is uninterpretable; with ``Market: Next president of South Korea?'', it becomes clearly \textsc{Pro}.

\paragraph{Design.} Four configurations are tested: $\{$3-class (\textsc{Pro/Anti/Neutral}), 2-class (\textsc{Pro/Anti} only)$\}$ $\times$ $\{$with context, without context$\}$. The 3-class vs.\ 2-class ablation tests whether including the dominant \textsc{Neutral} class (63.1\%) helps or hurts minority-class detection. The context ablation isolates the effect of market-question grounding. Input format with context: \texttt{"Market: \{question\} Comment: \{text\}"}.

\paragraph{Entity Masking.} To prevent the model from learning entity-specific shortcuts (e.g., memorizing that ``Trump'' co-occurs with \textsc{Pro} \citep{yuan2022debiasing}), NER-based entity masking is applied. SpaCy's \texttt{en\_core\_web\_sm} model identifies entities of types \texttt{PERSON}, \texttt{ORG}, \texttt{GPE}, \texttt{NORP}, \texttt{FAC}, \texttt{EVENT}, and \texttt{PRODUCT}; all matched spans are replaced with \texttt{ENTITY}. The token \texttt{ENTITY} is preferred over \texttt{[MASK]} because \texttt{[MASK]} carries special semantics from RoBERTa's masked language model pretraining and would send a conflicting signal to the encoder. Qualitative attention analysis confirming the effectiveness of this masking is shown in Figure~\ref{fig:attn_augmented}.

\subsection{LLM-Driven Counterfactual Augmentation}

With only 194 \textsc{Anti\_Position} samples (8.7\% of the data), the model is exposed to approximately 130 \textsc{Anti} examples during training, a number shown to be insufficient for learning a generalizable decision boundary even with class-weighted loss \citep{nakada2026synthetic}.

\paragraph{Motivation.} Generating linguistically similar \textsc{Pro}$\to$\textsc{Anti} flips creates grounded minority-class samples tied to specific market discussions, rather than generic synthetic text that would not match the underlying language distribution. This approach was pioneered for sentiment by \citet{kaushik2020learning} and adapted for stance by \citet{li2025factual}, but has not previously been applied to prediction market text.

\paragraph{Pipeline.} The Anthropic API (Claude Haiku 4.5, temperature 0.7, top-$p$ 0.9, max 150 output tokens) is prompted to flip $\sim$200 \textsc{Pro\_Position} comments to \textsc{Anti\_Position} while preserving trader tone, slang, and approximate length. The prompt includes the market question as context. The system prompt constrains output to raw comment text only, with no explanatory preamble.

\begin{figure}[H]
\centering
\includegraphics[width=0.95\linewidth]{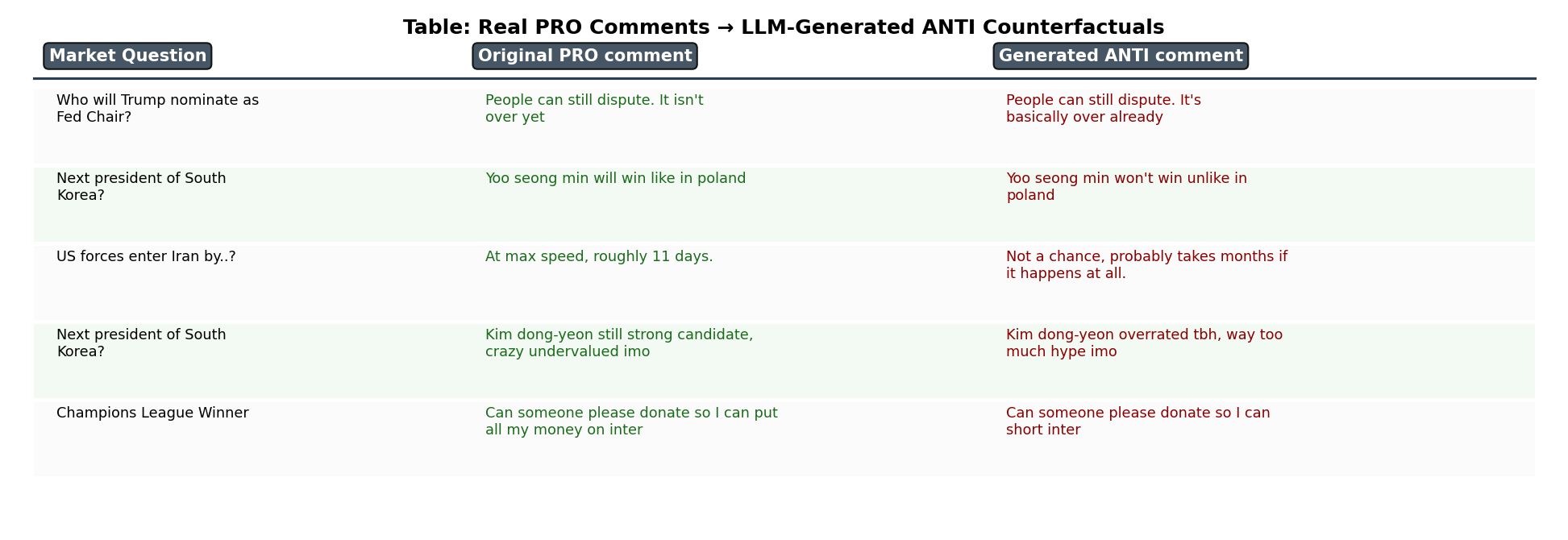}
\caption{Examples of counterfactual generation: original \textsc{Pro} comments and their LLM-generated \textsc{Anti} counterparts, with market question context.}
\label{fig:examples_table}
\end{figure}

Generated samples pass through automated quality filters: length-ratio bounds ($\leq 2\times$ original), token-overlap echo detection ($\leq 80\%$ overlap), meta-commentary removal (reject outputs beginning with ``Here'', ``Sure'', or ``I''), and minimum-length filtering ($\geq 3$ words). Two dose levels (50\% and 100\% of generated samples) are then evaluated, added to the training split only; validation and test splits remain 100\% real data throughout. Figure~\ref{fig:examples_table} shows representative generation examples.

\subsection{Attention-Based Interpretability}

The augmentation results reveal a non-obvious pattern (helping weak configurations but hurting strong ones) that aggregate metrics alone cannot explain.

\paragraph{Motivation.} Attention maps reveal which input tokens the model relies on for predictions, enabling mechanistic tests of why context helps and why augmentation has conditional effects. If augmented models shift attention away from stance-bearing tokens toward synthetic-pattern artifacts, this provides a direct explanation for performance degradation.

\paragraph{Design.} Last-layer attention weights over the \texttt{[CLS]} token are extracted and averaged across all 12 attention heads. Test examples are selected where baseline and augmented models disagree on their predictions, and attention distributions for the same input are compared across models. For the context ablation, identical comments are visualized with and without the prepended market question. Attention weights are treated as correlational rather than causal \citep{jain2019attention}, and are interpreted accordingly.

\section{Experimental Setup}
\label{sec:setup}

\subsection{Model Architecture}

\textbf{RoBERTa-base} \citep{liu2019roberta} (125M parameters, 12 transformer layers, 768 hidden dimensions) is employed via HuggingFace Transformers as the stance classifier. The classification head is a single linear layer mapping the \texttt{[CLS]} representation to the output classes: Linear(768 $\to$ $C$), where $C \in \{2, 3\}$.

To mitigate class imbalance, weighted cross-entropy loss is used with inverse class-frequency weights:
\begin{equation}
  w_c = \frac{N}{C \cdot N_c}
\end{equation}
where $N$ is the total number of samples, $C$ is the number of classes, and $N_c$ is the count of class $c$. This yields weights of 1.18 (\textsc{Pro}), 3.82 (\textsc{Anti}), and 0.53 (\textsc{Neutral}) for the 3-class setup.

\paragraph{Training hyperparameters.} AdamW optimizer \citep{loshchilov2019adamw} (weight decay = 0.01), learning rate $2 \times 10^{-5}$, batch size 16, maximum token length 128, gradient clipping (max norm = 1.0), maximum 10 epochs with early stopping on validation macro F1 (patience = 3).

\subsection{Dataset}

The dataset comprises 2,229 human-annotated comments from Polymarket, collected via the Polymarket Gamma API and CLOB API across 12 markets spanning three domains: politics (5 markets), sports (5 markets), and finance (2 markets). Each comment is annotated with \texttt{comment\_direction} (\textsc{Pro/Anti/Neutral}).

\begin{table}[h]
\centering
\caption{Dataset class distribution.}
\label{tab:distribution}
\begin{tabular}{lrr}
\toprule
\textbf{Class} & \textbf{Count} & \textbf{\%} \\
\midrule
\textsc{Neutral} & 1,407 & 63.1\% \\
\textsc{Pro\_Position} & 628 & 28.2\% \\
\textsc{Anti\_Position} & 194 & 8.7\% \\
\midrule
\textbf{Total} & \textbf{2,229} & \textbf{100\%} \\
\bottomrule
\end{tabular}
\end{table}

Data is split via stratified sampling: 3-class uses 1,559 / 335 / 335 (train/val/test); 2-class (dropping \textsc{Neutral}) uses 574 / 124 / 124. In augmented conditions, 108 (50\%) or 216 (100\%) synthetic \textsc{Anti} samples are added to the training split only.

\begin{figure}[H]
\centering
\includegraphics[width=0.95\linewidth]{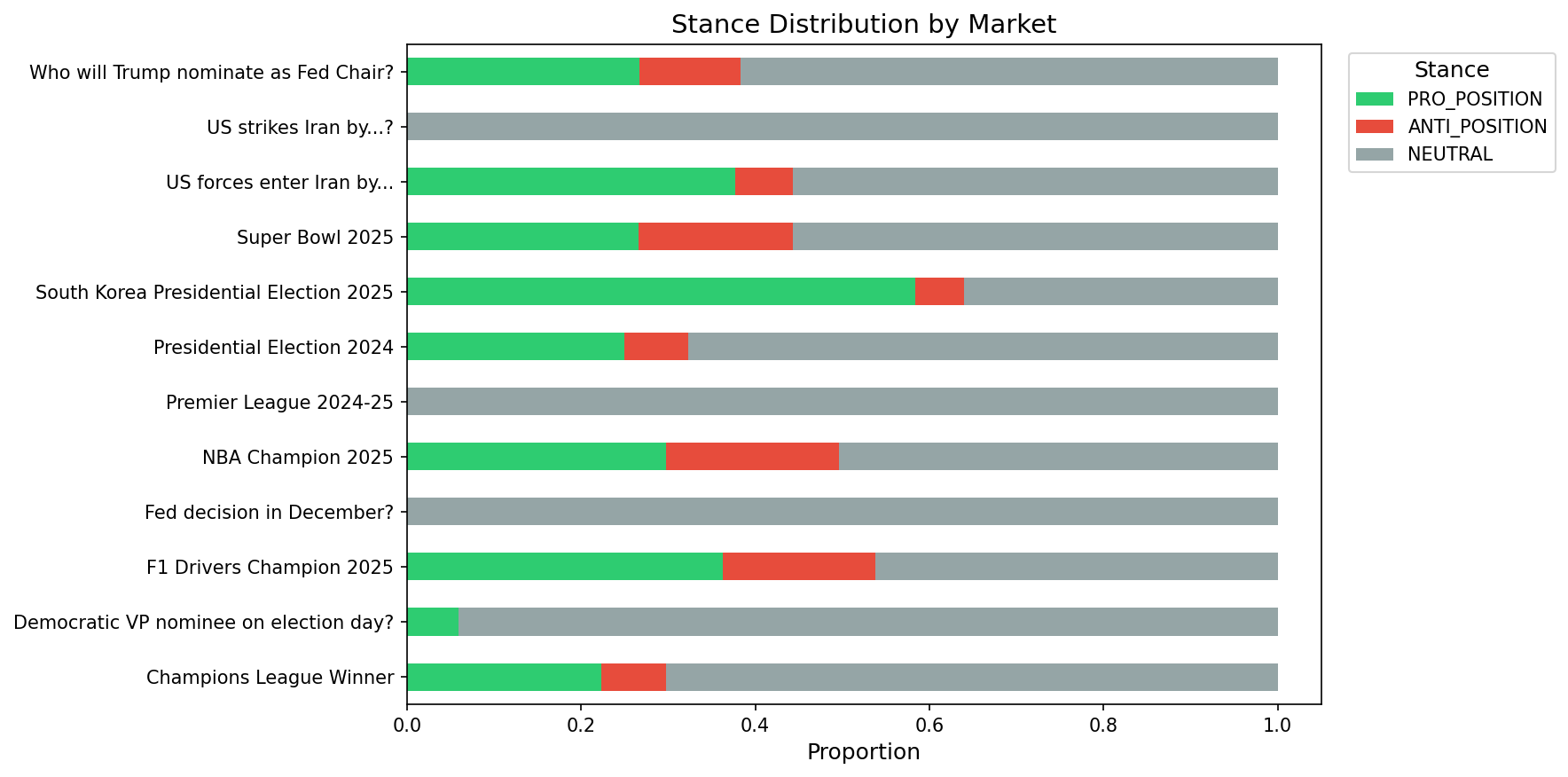}
\caption{Stance distribution by market. Stacked horizontal bar chart showing \textsc{Pro/Anti/Neutral} proportions per market across domains.}
\label{fig:stance_dist}
\end{figure}

\begin{figure}[H]
\centering
\includegraphics[width=0.85\linewidth]{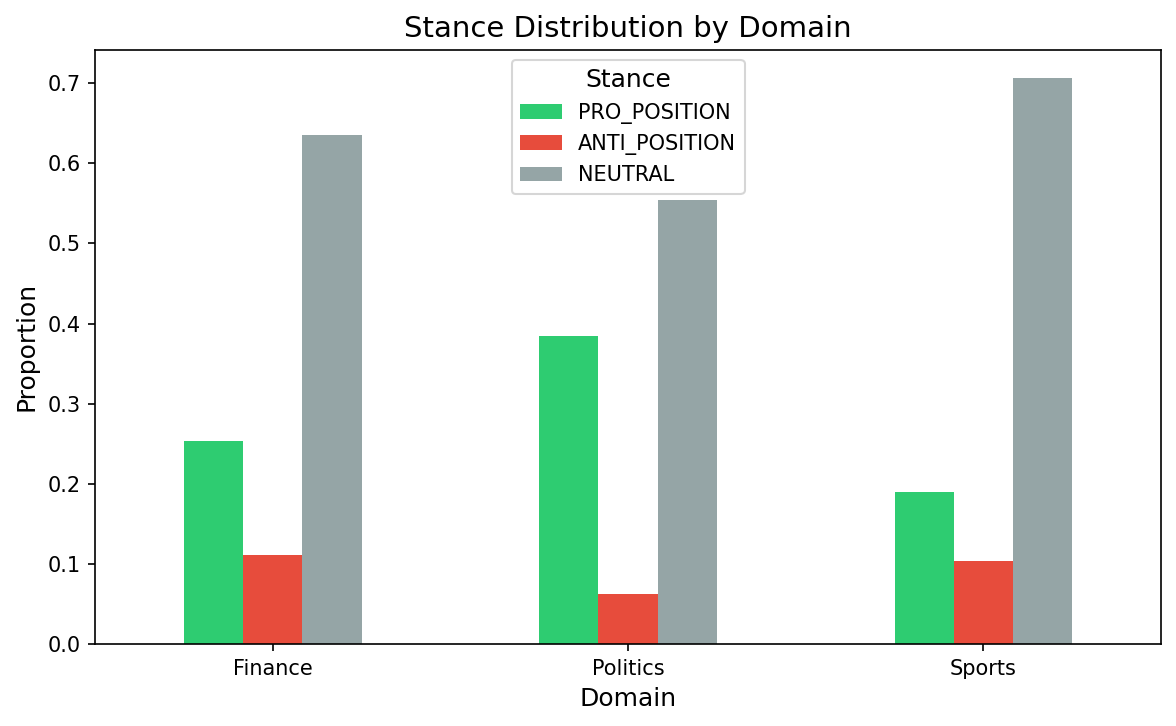}
\caption{Stance distribution aggregated by domain (politics, sports, finance). Shows cross-domain variation in directional commentary rates.}
\label{fig:stance_domain}
\end{figure}

\subsection{Evaluation Metrics}

\textbf{Macro F1} (primary metric) is the unweighted average of per-class F1 scores:
\begin{equation}
  \text{Macro F1} = \frac{1}{C} \sum_{c=1}^{C} F1_c = \frac{1}{C} \sum_{c=1}^{C} \frac{2 \cdot P_c \cdot R_c}{P_c + R_c}
\end{equation}
Macro F1 is critical in this setting because a \textsc{Neutral}-only classifier achieves 63\% accuracy but only $\sim$21\% macro F1 for 3-class. Per-class precision, recall, and F1 are additionally reported, with particular focus on \textsc{Anti} recall and \textsc{Anti} F1 as the targeted metrics for augmentation effectiveness. Accuracy is reported for completeness.

\section{Results}
\label{sec:results}

\subsection{Baseline Results}

Table~\ref{tab:baseline} presents baseline results across the four configurations. Market context is the dominant factor: in the 3-class setup, adding the market question increases \textsc{Anti} recall from 0.10 to 0.45 and macro F1 from 0.42 to 0.54. The best overall performance is achieved by the 2-class configuration with context (macro F1 = 0.68), benefiting from both the removal of the dominant \textsc{Neutral} class and the grounding provided by the market question.

\begin{table}[h]
\centering
\caption{Baseline results across four configurations. Best values in \textbf{bold}.}
\label{tab:baseline}
\begin{tabular}{lccccc}
\toprule
\textbf{Config} & \textbf{Macro F1} & \textbf{Anti P} & \textbf{Anti R} & \textbf{Anti F1} & \textbf{Acc.} \\
\midrule
3-class         & 0.42 & 0.10 & 0.10 & 0.10 & 0.65 \\
3-class + ctx   & 0.54 & 0.25 & 0.45 & 0.32 & 0.65 \\
2-class         & 0.54 & 0.29 & 0.48 & 0.36 & 0.60 \\
2-class + ctx   & \textbf{0.68} & \textbf{0.45} & \textbf{0.76} & \textbf{0.56} & \textbf{0.73} \\
\bottomrule
\end{tabular}
\end{table}

\subsection{Augmentation Ablation}

\begin{figure}[H]
\centering
\includegraphics[width=0.85\linewidth]{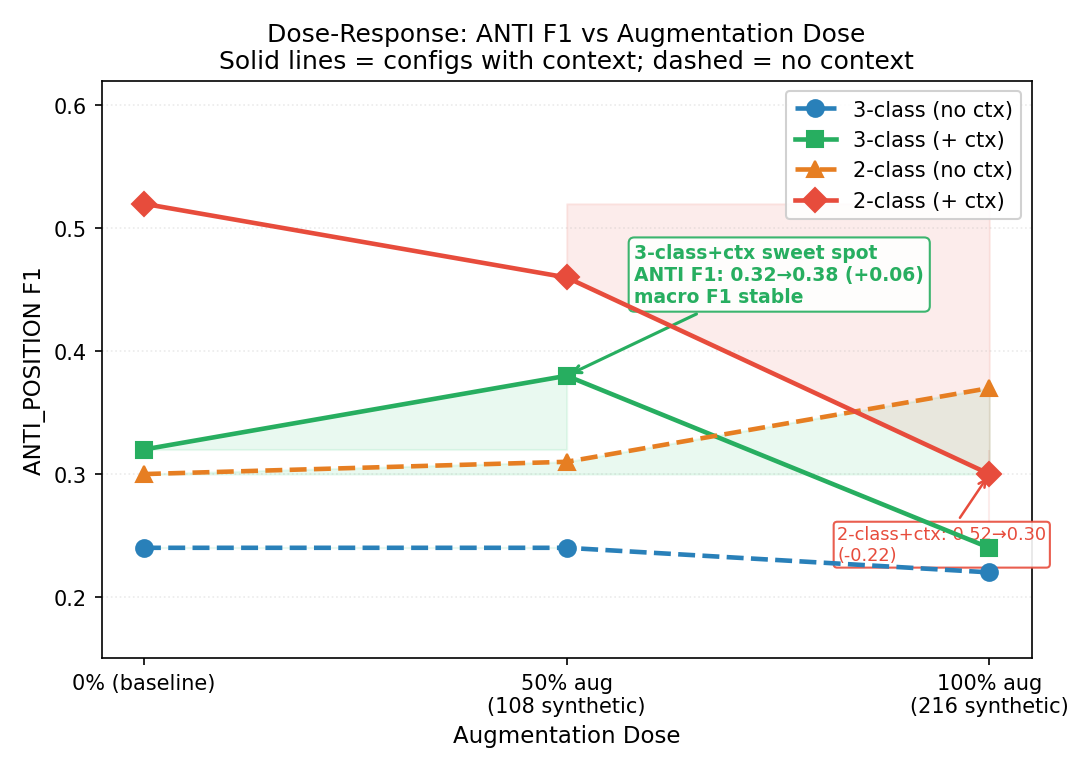}
\caption{Dose-response: \textsc{Anti} F1 vs.\ augmentation dose (0\%, 50\%, 100\%), one line per configuration. 3-class configurations exhibit an inverted-U while 2-class-ctx declines monotonically.}
\label{fig:dose_response}
\end{figure}

\begin{figure}[H]
\centering
\includegraphics[width=0.85\linewidth]{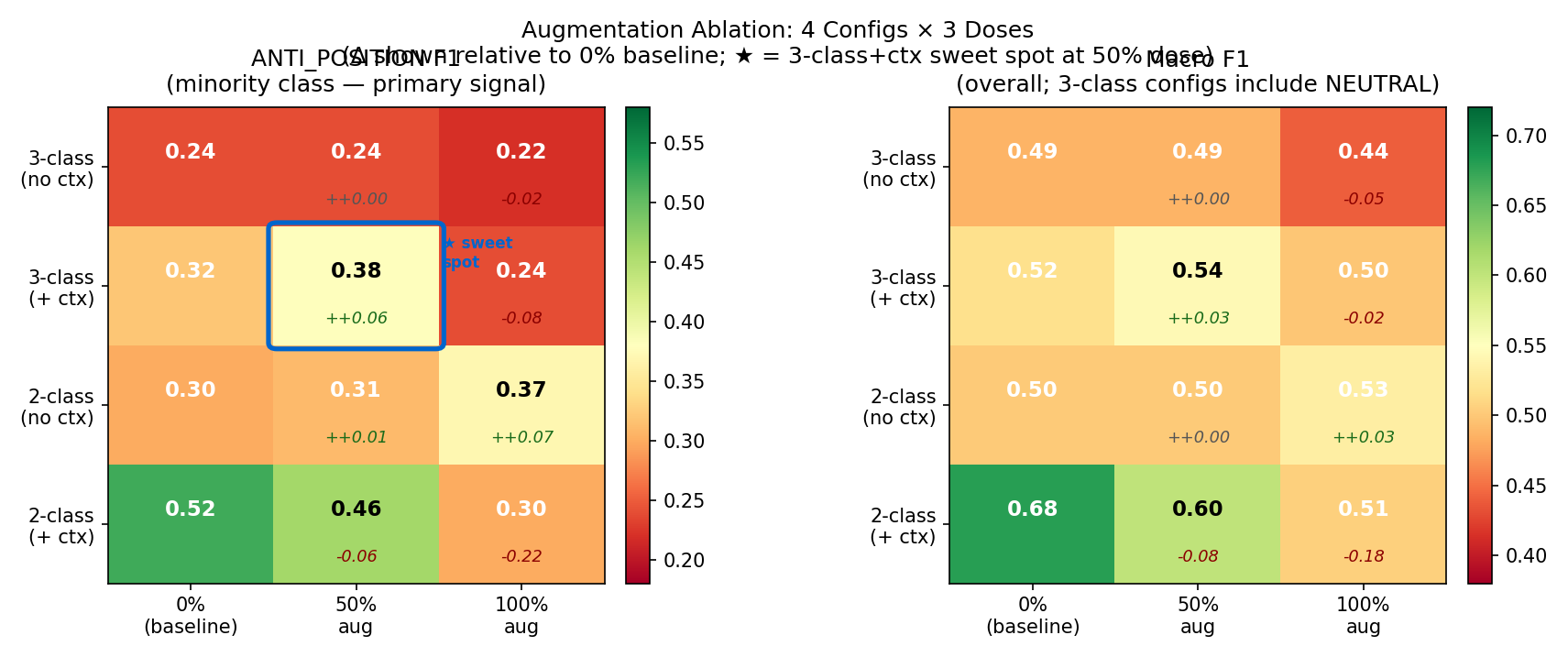}
\caption{Ablation heatmap: 4 configurations $\times$ 3 augmentation doses, cells color-coded by macro F1.}
\label{fig:heatmap}
\end{figure}

Table~\ref{tab:ablation} presents the full $4 \times 3$ ablation. Figures~\ref{fig:dose_response} and~\ref{fig:heatmap} visualize the dose-response and configuration-level trends respectively.

\begin{table}[h]
\centering
\caption{Full ablation: Macro F1 and Anti F1 across configurations and augmentation doses. Best per-config values in \textbf{bold}.}
\label{tab:ablation}
\begin{tabular}{ll cc cc cc}
\toprule
& & \multicolumn{2}{c}{\textbf{Baseline}} & \multicolumn{2}{c}{\textbf{50\% Syn}} & \multicolumn{2}{c}{\textbf{100\% Syn}} \\
\cmidrule(lr){3-4} \cmidrule(lr){5-6} \cmidrule(lr){7-8}
\textbf{Config} & & M-F1 & A-F1 & M-F1 & A-F1 & M-F1 & A-F1 \\
\midrule
3-class       & & 0.42 & 0.10 & \textbf{0.49} & \textbf{0.24} & 0.44 & 0.22 \\
3-class + ctx & & \textbf{0.54} & 0.32 & \textbf{0.54} & \textbf{0.38} & 0.50 & 0.24 \\
2-class       & & \textbf{0.54} & \textbf{0.36} & 0.50 & 0.31 & 0.53 & 0.37 \\
2-class + ctx & & \textbf{0.68} & \textbf{0.56} & 0.60 & 0.46 & 0.50 & 0.30 \\
\bottomrule
\end{tabular}
\end{table}

Three key patterns emerge from the ablation:

\begin{enumerate}[nosep]
  \item \textbf{50\% augmentation is the optimal dose} for 3-class configurations: 3-class macro F1 improves from 0.42 to 0.49, and 3-class-ctx achieves its best \textsc{Anti} F1 (0.38) with no loss in overall macro F1.
  \item \textbf{100\% augmentation degrades performance}, most severely for 2-class-ctx (macro F1 drops from 0.68 to 0.50, a 26\% relative decrease).
  \item \textbf{2-class configurations are more sensitive to augmentation noise} than 3-class, as every synthetic sample directly competes on the \textsc{Pro}/\textsc{Anti} boundary without a \textsc{Neutral} buffer.
\end{enumerate}

\begin{figure}[H]
\centering
\includegraphics[width=0.75\linewidth]{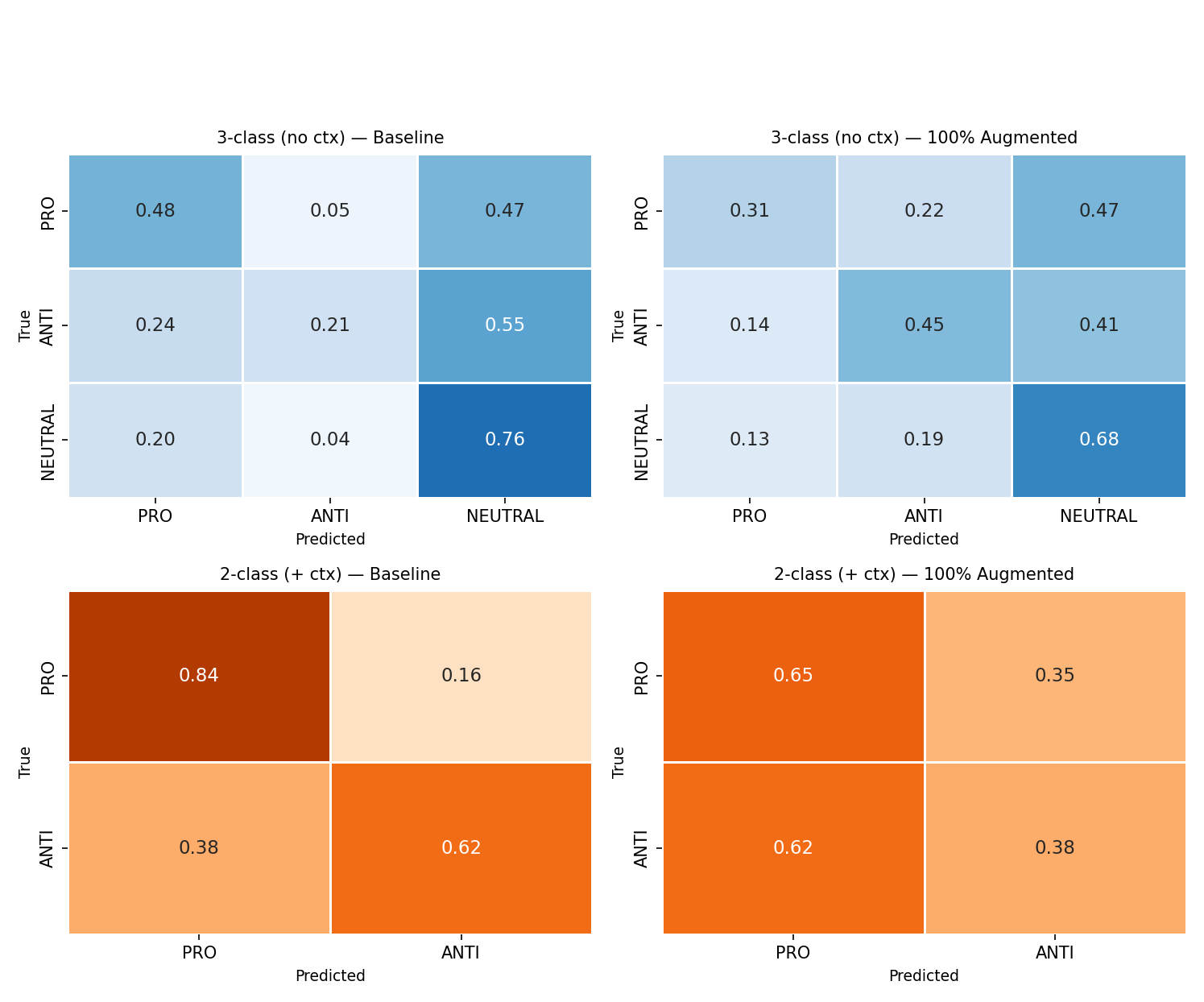}
\caption{Confusion matrices (row-normalized). Top row: 3-class baseline vs.\ 3-class 100\% augmented. Bottom row: 2-class-ctx baseline vs.\ 2-class-ctx 100\% augmented.}
\label{fig:confusion}
\end{figure}

\begin{figure}[H]
\centering
\includegraphics[width=0.75\linewidth]{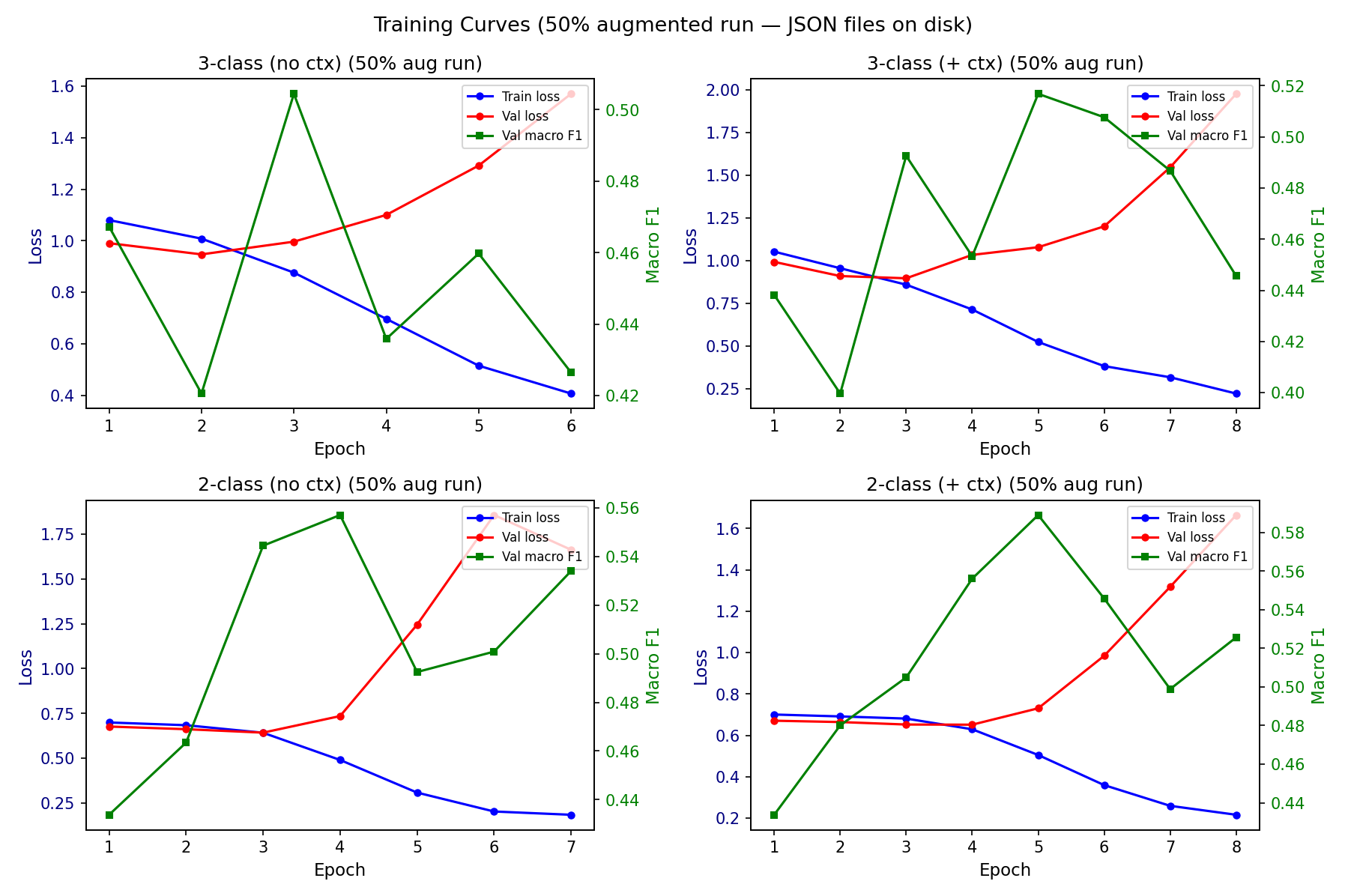}
\caption{Training curves: validation macro F1 per epoch for baseline vs.\ 50\% augmented vs.\ 100\% augmented. Left: 3-class-ctx. Right: 2-class-ctx.}
\label{fig:training_curves}
\end{figure}

\section{Analysis and Discussion}
\label{sec:analysis}

Three primary failure modes are identified, with mechanistic hypotheses provided for each.

\paragraph{Failure Type 1.} \textsc{Anti} comments misclassified as \textsc{Neutral}: short, low-signal opposition (e.g., ``nah'', ``doubt it'') lacks sufficient stance-bearing language for the model to distinguish from neutrality.

\paragraph{Failure Type 2.} \textsc{Anti} comments misclassified as \textsc{Pro}: sarcastic or ironic comments where surface sentiment is positive but stance is negative (e.g., ``yeah sure, this is definitely happening lol'') \citep{joshi2017sarcasm}.

\paragraph{Failure Type 3.} \textsc{Pro} comments misclassified as \textsc{Anti} after augmentation: synthetic \textsc{Anti} patterns, which tend to be longer and more grammatically complete than real trader opposition, leak into the model's representation of real \textsc{Pro} comments.

\subsection{Hypothesis 1: Context Resolves Ambiguous Short Comments}

\begin{figure}[H]
\centering
\includegraphics[width=0.85\linewidth]{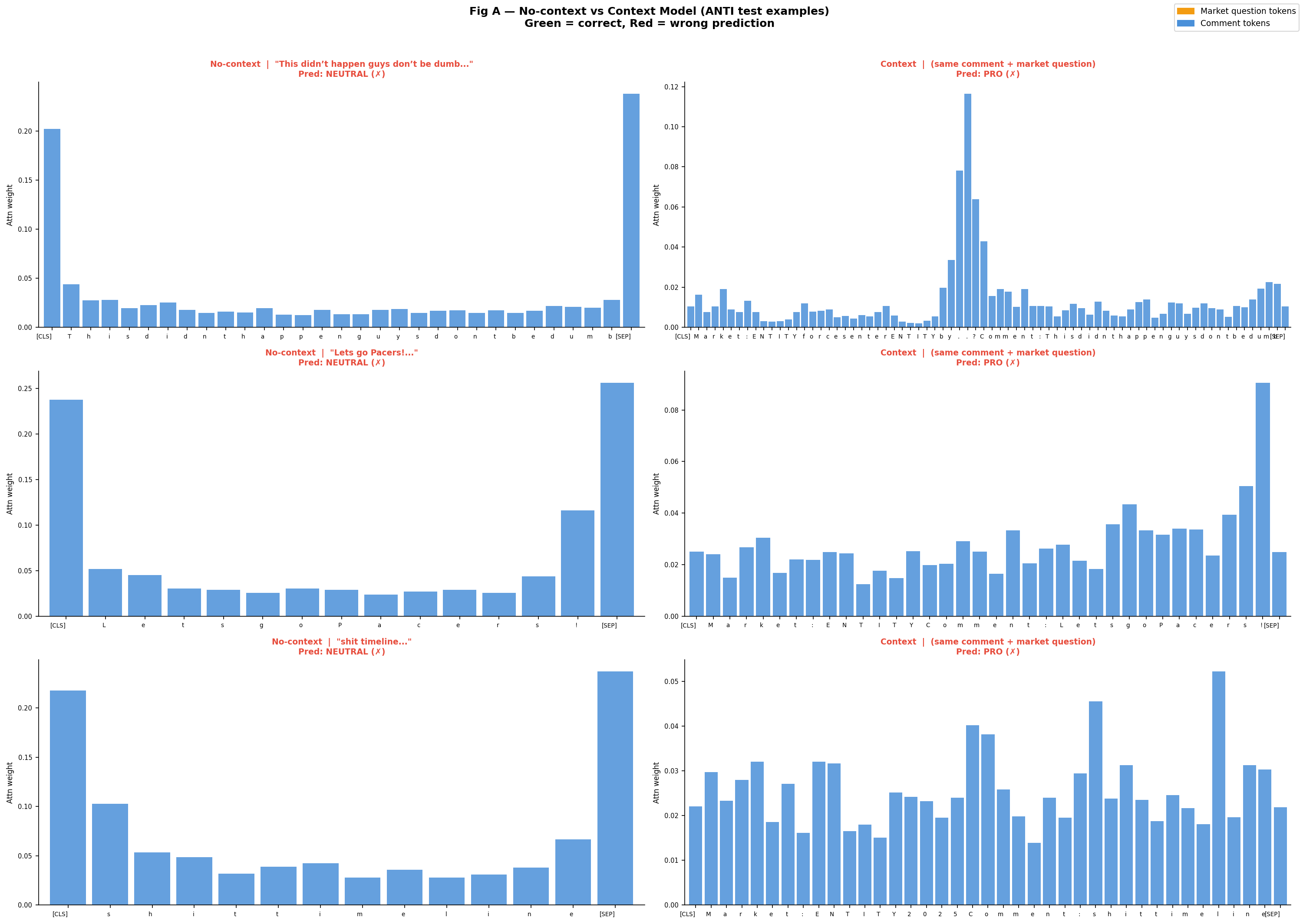}
\caption{Attention maps: with (right) vs.\ without (left) context on the same \textsc{Anti} comment. Context enables the model to attend to market question tokens for stance disambiguation, producing a more even distribution of attention across the statement.}
\label{fig:attn_context}
\end{figure}

Without the market question, comments such as ``this is moving'' are ambiguous, as the model cannot determine what ``this'' refers to or whether ``moving'' implies a favorable or unfavorable outcome. As shown in Figure~\ref{fig:attn_context}, without context the model attends predominantly to the start and end tokens (a proxy for text length \citep{clark2019bert}), while with context, attention is distributed across substantive tokens in both the market question and the comment body. This distributional shift is consistent with the macro-level confusion matrix improvements shown in Figure~\ref{fig:confusion}.

\subsection{Hypothesis 2: Augmentation Introduces Distribution Mismatch at High Doses}

LLM-generated \textsc{Anti} comments do not faithfully replicate the stylistic properties of real Polymarket opposition. Real \textsc{Anti} comments are terse and slang-heavy (``this is cooked'', ``nah rip''), while synthetic comments tend to be longer and more grammatically complete. Table~\ref{tab:examples} illustrates representative failure cases in generation quality.

\begin{table}[h]
\centering
\caption{Real vs.\ synthetic \textsc{Anti} examples selected for poor generation quality: loss of trader prose (2nd example) and semantic incoherence (1st and 3rd examples).}
\label{tab:examples}
\begin{tabular}{p{3cm}p{4.5cm}p{4.5cm}}
\toprule
\textbf{Market} & \textbf{Original Pro} & \textbf{Generated Anti} \\
\midrule
\textit{Presidential Election Winner 2024?} & ``You lost get over it trump male everything better'' & ``You won but trump didn't make everything better'' \\
\textit{US forces enter Iran by..?} & ``The kurds are going in too. The stories are now seemingly all allowed to come out. UK media is talking about UK guys joining the kurds for the ground invasion.'' & ``The kurds aren't going in. The stories being suppressed suggest nothing's actually happening. UK media is totally silent about UK guys joining any ground ops, this whole invasion narrative is overblown.'' \\
\textit{Who will Trump nominate as Fed Chair?} & ``To the victor goes the spoils'' & ``To the loser goes the spoils'' \\
\bottomrule
\end{tabular}
\end{table}

This distribution mismatch hypothesis predicts the observed conditional pattern: when the baseline already achieves strong \textsc{Anti} recall (2-class-ctx: 0.76), injecting distribution-shifted synthetic data corrupts the decision boundary. When baseline \textsc{Anti} recall is near-zero (3-class: 0.10), even noisy synthetic signal provides useful boundary information.

\subsection{Hypothesis 3: Market-Level Variation in Anti Detection}

Markets characterized by implicit or ironic commentary may exhibit lower \textsc{Anti} recall than markets with direct oppositional language. Sports markets tend toward explicit opposition (``no way they win''), while political and financial markets rely more heavily on irony and implication. This is reflected in the data distributions shown in Figures~\ref{fig:stance_domain} and~\ref{fig:stance_dist}: several markets (``Fed decision in December?'', ``Democratic VP nominee...'', ``US Strikes on Iran by...'') contribute zero \textsc{Anti} comments to the dataset, limiting the model's exposure to domain-specific opposition patterns.

\section{Conclusion}
\label{sec:conclusion}

This paper presented the first stance detection study targeting prediction market commentary, a domain distinguished by extreme brevity, trader vernacular, and severe class imbalance. Across a 12-run ablation, three principal findings are established. First, market context is the dominant performance driver: prepending the market question to each comment raises 3-class \textsc{Anti} recall from 0.10 to 0.45 and achieves a best overall macro F1 of 0.68 in the 2-class setting, a free input transformation that requires no architectural changes. Second, LLM-driven counterfactual augmentation is conditionally beneficial: it substantially improves weak configurations but degrades strong ones, with the direction of effect determined by the gap between the baseline's \textsc{Anti} recall and the distributional fidelity of the synthetic samples. Third, 50\% augmentation consistently outperforms 100\%, confirming domain-specific diminishing returns consistent with prior work \citep{karimi2023caisa}.

These results carry practical implications for deploying stance classifiers on financial social media. Market context should always be included as an input feature. Counterfactual augmentation should be applied cautiously and at a conservative dose, and its benefit should be validated against a held-out real-data split before deployment.

Future work should address three open challenges: (1) improving synthetic sample fidelity by fine-tuning the generation model on real \textsc{Anti} examples before augmentation; (2) extending the dataset to cover more markets and a broader range of opposition styles, particularly in finance; and (3) developing sarcasm-aware representations \citep{joshi2017sarcasm} to address Failure Type 2, which constitutes a structurally distinct misclassification regime not amenable to augmentation-based solutions.

\bibliographystyle{plainnat}

\appendix
\end{document}